# Modular Controllers Facilitate the Co-Optimization of Morphology and Control in Soft Robots

Alican Mertan
alican.mertan@uvm.edu
Neurobotics Lab
University of Vermont
Burlington, VT, USA

Nick Cheney
ncheney@uvm.edu
Neurobotics Lab
University of Vermont
Burlington, VT, USA

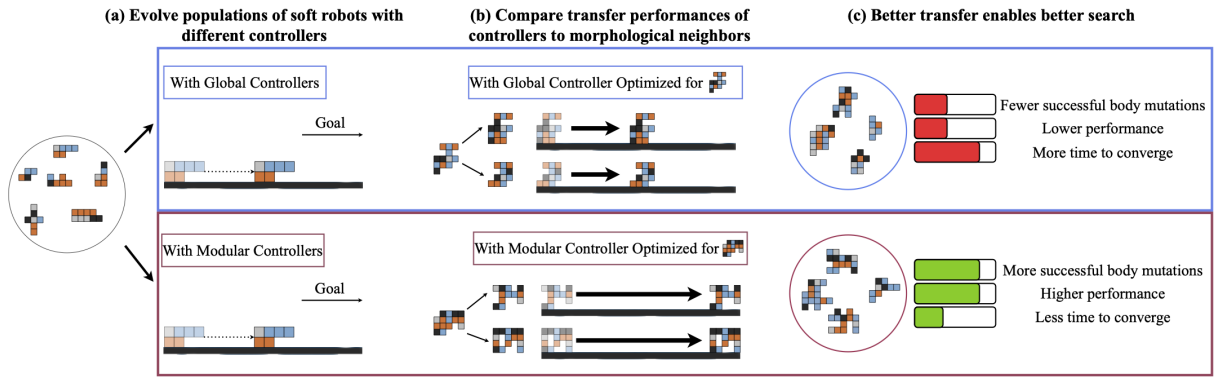

Figure 1: Main hypothesis: In the brain-body co-optimization of soft robots, the locomotion performance of modular controllers is significantly more robust than global controllers to perturbations of their robot's morphology. This allows better transfer of that controller to a robot's offspring with morphological mutations. This increases the rate of successful morphological mutations during the co-optimization process and leads to a more efficient and effective search over robot designs.

## ABSTRACT

Soft robotics is a rapidly growing area of robotics research that would benefit greatly from design automation, given the challenges of manually engineering complex, compliant, and generally non-intuitive robot body plans and behaviors. It has been suggested that a major hurdle currently limiting soft robot brain-body co-optimization is the fragile specialization between a robot's controller and the particular body plan it controls, resulting in premature convergence. Here we posit that modular controllers are more robust to changes to a robot's body plan. We demonstrate a decreased reduction in locomotion performance after morphological mutations to soft robots with modular controllers, relative to those with similar global controllers – leading to fitter offspring. Moreover, we show that the increased transferability of modular controllers to similar body plans enables more effective brain-body co-optimization of soft robots, resulting in an increased rate of positive morphological mutations and higher overall performance of evolved robots. We hope that this work helps provide specific methods to improve soft robot design automation in this particular setting, while also providing evidence to support our understanding of the challenges of brain-body co-optimization more generally. [1]

## 1 INTRODUCTION

A highly touted feature in the evolution of biological creatures is the adaptive advantage of complex biological forms and the environmental and behavioral niches which these unique morphologies satisfy [13, 14, 46, 56]. Artificial creatures and engineered robots also benefit from highly effective and efficient body plans [41]. This is especially true in soft robots, where flexible and compliant materials enable a variety of complex robots, and lifelike behaviors that are inaccessible to their rigid counterparts [25, 48, 51]. These materials also hold an increased potential for morphological computation [41, 42], making the design of robot body plans, and the tight integration of that body plan with its control strategy, especially critical to the robot's performance.

---

[1]Code repository: https://github.com/mertan-a/gecco-23

However, these advantages also come with a significant challenge: the compliance, complexity, and abundance of non-linear interactions across the form and dynamics of a soft robot lead to a particularly non-intuitive design space, suggesting the increased potential for automated design over manual engineering. Despite the interest and potential in brain-body design automation, co-optimization of morphology and control through evolutionary algorithms remains an open and challenging problem. It has been suggested that this is due, in part, to the specialization of robot controllers and behaviors to the particular morphologies that they inhabit – as tightly coupled and specialized controllers or body plans are not amenable to change in either component without a coordinated change in the other [6].

While prior work attempted to create genetic representations that increase the likelihood of more coordinated changes [54, 58] or rely on diversity maintenance to reduce selection pressure during the re-adaption of brain and body post-mutation [7, 31], all of these works accept the fragile co-adaptation at the heart of the problem as a given. Here, we take a slightly different perspective and ask how we can reduce the amount of fragility in our co-optimization by making the components of our brain-body system more robust to changes in the other – which, we hypothesize, will help to reduce the challenges of making successful mutations to only one component of the system and enable more effective brain-body co-optimization. In effect, we are suggesting that certain implementation decisions about our morphologies and controllers may smooth the fitness landscape by reducing the coupling between the controller and morphological genes/parameters.

This philosophy could be studied in the context of producing more robust morphologies, though we first start in this work focusing on an investigation of more robust controllers to morphological changes. In particular, we note the robustness attributed to modular controllers [63] and firstly hypothesize that soft robots undergoing brain-body co-optimization with modular controllers will more effectively transfer to offspring with morphological mutations, then secondarily hypothesize that an increased rate of positive morphological mutations will lead to an overall increase in the effectiveness of brain-body co-optimization in soft robots.

## 2 RELATED WORK

**Modular Control** Modularity is considered an important feature and is under investigation in both natural [60] and artificial systems [9]. Especially for robotics, modular robots are considered versatile, robust, and adaptive [63]. Yet it is challenging to design and control such systems, and it is an active area of research in both rigid [22, 40, 61] and soft robotics [7, 8, 36, 43].

**Soft Robotics** The field of soft robots with volumetric actuation started with [19, 21, 55] and with the availability of simulators such as [3, 20, 32, 35], many others have followed. Soft robots are evolved for locomotion tasks in different environments [8, 11, 24, 28], their ability to change their shape volumetrically are investigated [5, 29, 50], different types of control strategies are developed [12, 17, 34, 43]. Lifetime development in a co-optimization setting is studied in [10, 26, 27]. The difficulty of co-optimization due to fragile co-dependence of brain and body is explored [6], and algorithmic solutions that combat the resulting premature convergence through increased diversity are proposed [7]. Different representations and their effects on the evolutionary optimization process are studied in [33, 44, 54, 57, 58].

Closer to our work are the works of [22, 30, 34, 36, 43]. Huang et al. [22] train modular controllers with a message-passing scheme to control rigid robots for a locomotion task. Medvet et al. [34] evolve modular controllers with message-passing for various fixed morphologies. In follow-up work, Medvet et al. [36] co-optimizes morphology and control for soft robots but focuses on the effects of evolutionary algorithm and representation on biodiversity and performance. Pigozzi et al. [43] evolves modular controllers that use indexing and self-attention mechanism for soft robots with fixed morphologies. Kvalsund et. al. [30] explore centralized and decentralized control in modular rigid robots and demonstrate the trade-off between them. As opposed to [22, 34, 43], our modular controllers don't use any message-passing scheme or indexing and can work with arbitrary morphology without any change. Instead of experimenting with fixed morphologies as in [22, 34, 43], we focus on the more challenging problem of co-optimization of morphology and control and investigate the dynamics of co-optimization with modular controllers.

## 3 METHODS

### 3.1 Simulation

Our work uses the open-source Evolution GYM (EvoGym) benchmark [3]. It consists of a mass-spring system-based soft-body simulation engine and various task environments. Similar to the simulation engines in [17, 24, 34–37, 43], EvoGym works in 2D. The simulation engine and the provided environments are open-source and provide Python API for fast prototyping and experimenting. Please see Section 4 for the details of the environment used in our experiments.

The soft robots are represented as a mass-spring system in a grid-like layout. Each voxel is initialized as a cross-braced square with masses in four corners and ideal springs in the edges. These springs can have different spring constants depending on the voxel material type. A voxel can be initialized from rigid or soft passive material or horizontally or vertically actuating active material. Figure 4 shows example robots with all four materials. The black voxels are rigid, the grey voxels are soft, and the orange and blue voxels are horizontal and vertical actuating voxels, respectively. The active materials' color shade represents their volume and gets darker as the voxel contracts and gets lighter as the voxel expands. Following the standard practice of using a bounding box in the literature [6, 7, 12, 26, 33, 36, 54], we limit the robot design to a $5x5$ bounding box in our experiments to keep the design space tractable.

The simulation provides a number of observations coming from the robot and the environment. A controller can observe a voxel's velocity $V \in \mathbb{R}^2$, its volume $v \in \mathbb{R}$, and its material (or its absence) as a one-hot encoded vector $M \in [0, 1]^5$. A periodic time signal $t \in [0..24]$ (simulation time step mod 25) is also available to help controllers to create a periodic behavior.

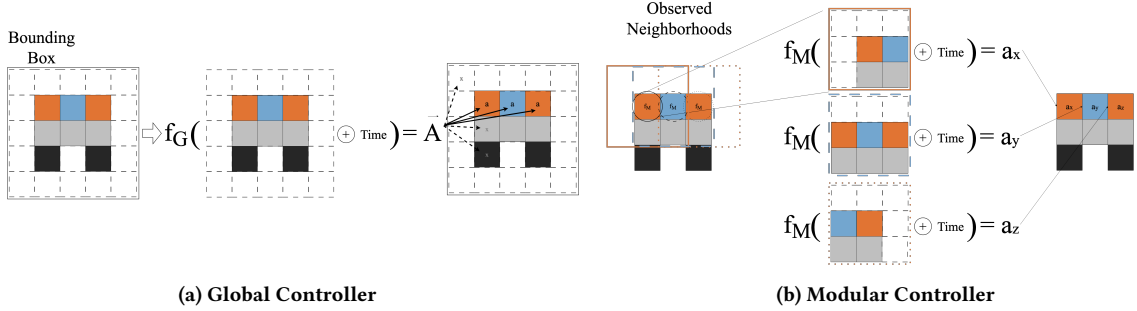

(a) Global Controller          (b) Modular Controller

Figure 2: Illustration of controller variants: (a) The global controller utilizes all the available information and outputs an action matrix that contains actions for all possible voxels. Actions for active voxels are assigned, and the rest is discarded. (b) Each active voxel shares the same modular controller. Observations from a local neighborhood are used to predict actions. We use the observation range of 1 in the figure for ease of illustration, and a Moore neighborhood of size 2 in our experiments.

## 3.2 Control paradigm

The soft robots are controlled by specifying the actuation ($a \in [0, 1]$) of each voxel with active material in the robot at each time step of the simulation. The controllers are modeled with a function $f$ that maps observations $O$ to actions $A$ as $f(O) = A$.

**Global Control** Similar to [17, 22, 34], we devise a straightforward control for the robots where a centralized, global controller processes the complete information of the soft robot to assign actions to each voxel, as illustrated in Figure 2a. Observations coming from each voxel are concatenated into vector $O$, and the global controller $f_G$ processes all the available information to output vector $A$ that contains actions for each voxel as in $f_G(O) = A$.

Given this formulation of global controllers and the brain-body co-optimization setting, compatibility problem could arise. A controller $f_G$ designed for a particular morphology could be incompatible with another morphology that comes up during optimization, which causes issues for simultaneous optimization of control and morphology. To overcome this issue and create a fair comparison with modular control, we use a simple caching trick similar to the one used in [22], where the global controllers always expect observations from and output actions for the biggest robot possible (a 5x5 robot). To make it compatible with any given morphology in our design space, we zero-pad the observations for missing voxels and mask out the unnecessary actions for the morphology at hand.

**Modular Control** We investigate the modular control of soft robots due to voxel-based soft robots' aptness for modular control and the advantages of modularity [22, 34, 36, 43, 63]. In modular control, each active voxel of the soft robot contains a copy of the same controller $f_M$ with the same parameters $\theta_M$. As shown in Figure 2b, these controllers make observations from a local neighborhood of voxels $N^d$ with distance $d$ and output an action $a$ for the voxel they belong to. Therefore modular controller $f_M$ takes the form: $f_M(O_i) = a_i$, where $O_i$ is the observation for the $i^{th}$ voxel and consists of the concatenation of observations from each voxel belonging to Moore neighborhood $N_i^d$ of voxel $i$ with distance $d$, $a_i$ is the action for the $i^{th}$ voxel, and $i \in$ active voxels. If a voxel in the neighborhood $N_i^d$ is missing, its velocity and volume are observed as a zero-vector and zero, respectively. This allows the controller to make sense of its local neighborhood's structure. We also note that this formulation of modular control is analogous to neural cellular automatons [38].

By design, the modular controller is agnostic to the robot's morphology in the sense that it can work with any robot morphology without needing any change. Importantly, since each voxel shares the same controller, the difference in the behavior arises from the different observations each controller makes.

## 3.3 Controller model

Following the common practice of utilizing neural networks as powerful function approximators [17, 22, 34, 36, 37, 43, 44, 52–54], the controllers $f_M$ and $f_G$ are modeled by a single hidden layer MLP with learnable parameters $\theta_M$ and $\theta_G$, respectively. The hidden layer consists of 32 units with ReLU activations for both controllers and maps the observations into a feature vector. The MLP for modular controllers has a single output unit with a sigmoid activation, outputting an action based on the feature vector. The MLP for global controllers has 25 output units with sigmoid activations that map the feature vector to actions for each voxel separately.

In all of our experiments, we use a Moore neighborhood of distance $d = 2$ for modular controllers that we choose empirically based on our initial experiments and to have a similar number of parameters for both controllers. Additionally, we assume a 5x5 bounding box for the design space for global controllers. With these models and hyperparameters, both controllers have 201 inputs, modular controllers have 6497 parameters, and global controllers have 7289 parameters to optimize. While global controllers have slightly more parameters to optimize, they have the advantage of separately tuning the behavior of each voxel. On the other hand, modular controllers have fewer parameters to optimize, but changes in the controller could potentially affect the behavior of all voxels.

## 3.4 Training algorithm

Following [6, 7, 24, 36, 44, 54, 57, 58], we use an evolutionary algorithm to optimize the morphology and control of the soft robots simultaneously. In particular, we use $(\mu + \lambda)$-Evolution Strategies where $\mu = 16$ and $\lambda = 16$. Similar to [7, 26], we add a random individual to the population at each generation to increase diversity. The selection is based on a multi-objective Pareto ranking on an individual's age and fitness as in [49], where the individual's age is increased at every generation and set to 0 for newly created individuals. Unlike [49], we also set age to 0 after each mutation, incentivizing high levels of diversity and turnover in the population, loosely related to a less extreme version of Real et al. [47]'s method of giving all children a selection advantage over their parents. Recombination was not considered in this work.

The morphology is represented directly in the genome as a 2D matrix consisting of materials of voxels ([1..4]) or 0 if no voxel exists in that location. New morphologies are created through mutating existing morphologies or mutating empty morphology. Similar to [3], the mutation operator for morphology works by going through each possible voxel location and changing it randomly to one of [0..4] with 10% probability. We also ensure that each morphology has at least two active materials and 20% of their voxels filled by rejecting mutations that violate these constraints.

The controller genome consists of a vector of parameters ($\theta_M$ or $\theta_G$). New controllers are created from scratch by Pytorch's default initialization [39] or created through mutating existing controllers by adding a noise vector of the same size sampled from $\mathcal{N}(0, 0.1)$.

During evolution, offspring are created either by mutating an existing individual's controller or its morphology. Following [7], we heuristically choose 50% probability to decide which part of the individual to be mutated.

## 4 EXPERIMENTS

We use EvoGym's *Walker-v0* environment [3] as the locomotion task for the evolved robots. In this task, the robot must locomote in a flat terrain as far as possible. We use a modified reward function

$$R(r, T) = \Delta p_x^r + \mathbf{I}(r) + \sum_{t=0}^{T} -0.01 + 5, \quad (1)$$

to encourage the robot $r$ to move as fast as possible, where $\Delta p_x^r$ is the change in robot $r$'s position in the positive x direction, $\mathbf{I}(r)$ is the indicator function that takes the value of 1 if the robot $r$ has reached to the end of terrain and 0 otherwise. The summation term applies a small penalty at each time step to encourage the robot to reach the end of the terrain faster. The last term, a positive constant equal to the max time penalty, is used to shift the rewards to be positive for ease of analysis. The environment runs until the robot reaches the end of the terrain or 500 time steps, whichever happens first. Similar to [34, 36, 43], we apply action repetition to speed up the simulations and prevent controllers from exploiting high-frequency dynamics. The controllers are queried every fourth time step, and the last actions are repeated in other time steps.

We apply the Wilcoxon rank-sum test to report the statistical significance of our results wherever possible.

### 4.1 Brain-Body co-optimization

We compare the global controller and the modular controller in the problem of simultaneously optimizing morphology and control. We run the evolutionary algorithm described in Section 3.4 for 5000 generations for each controller and repeat the same experiments 100 times with different random seeds.

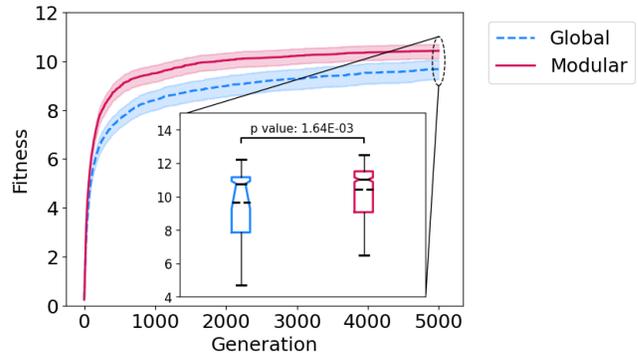

**Figure 3: Comparison of modular and global controllers on the problem of brain-body co-optimization for the experimented locomotion task. Modular controllers adapt more rapidly in the early generations and significantly outperform global controllers after 5000 generations of brain-body co-optimization. Lines show the mean, and shaded regions show the 95% confidence intervals calculated over 100 experiments. The boxplot shows the distribution of the performance of champions over the 100 trials – demonstrating higher mean, median, max, and min performance for modular controllers.**

Figure 3 illustrates the results of brain-body co-optimization experiments. In the main plot, we plot the performance of the best individuals that evolved in each generation. While the solid lines show the average fitness values of best-performing individuals over 100 repetitions, the shaded region shows the 95% bootstrapped confidence intervals. Both experimental treatments considered converged as more than 90% of the runs in each treatment are not showing improvement bigger than 0.1 in fitness for the last 500 generations at the 5000 generations mark. We find that the best solution found ("run champion") across each of the 100 trials is, on average, more fit for modular controllers (fitness of 10.42) than global controllers (9.67; p=0.0016). It is also the case that the champions evolved in runs featuring modular controllers displayed less variation than trial employing a global controller (range: [6.51, 12.5] vs. [4.73, 12.23]; IQR: [9.1, 11.5] vs. [7.9, 11.2] ) (inset plot in Figure 3). If robots with modular controllers were simply a scaled-up faster version of their globally controlled counterparts, we might expect the modular control treatment to have both higher average fitness and higher variability. This reduction in absolute variability across runs, despite higher overall values, may be suggestive of fundamentally differing fitness landscapes in the two settings – or perhaps differing abilities of evolution in these two settings to escape local optima in similarly rugged fitness landscapes.

In addition to reaching a higher level of fitness, robots evolved with modular controllers converge to their max fitness value (the fitness level found at generation 5000) significantly (all p<0.05) faster than those with global controllers – with modular controllers reaching 99% of their final performance by generation 2710.06 on average compared to 3059.37 for global controllers (and 1379.71 generations for modular vs. 1786.89 for global controllers to reach 95% of their respective max fitness levels, 1001.74 gens vs. 1466.28 to reach 90%, and 439.84 vs. 709.61 to reach 80%).

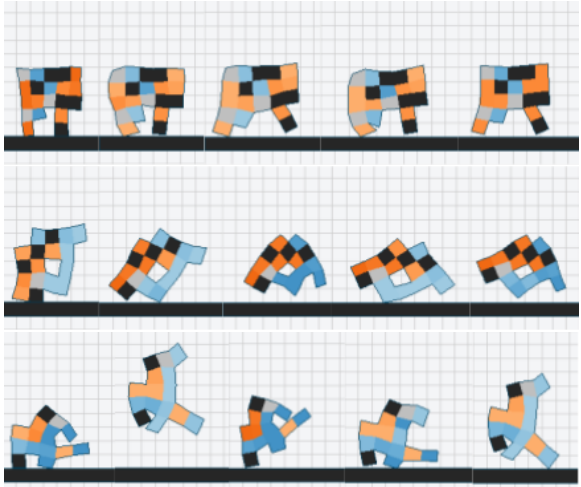

Figure 4: Time series of gaits, as robots move from left to right, exemplify robots evolved with modular control.

Figure 4 shows a few of the champions that evolved with modular controllers. The gaits are shown as a time series of snapshots. Similar to [8, 36], we see diverse shapes and behaviors evolve. While the robot at the top uses two leg-like limbs to gallop forward, the middle robot starts its movement by falling forward and then balances itself on a single leg that is used to throw itself forward. In the case of the bottom robot, we see more vertical movement where the robot jumps forward with the help of its forward limb and spends much of its time in the air. It exemplifies the ability to evolve diverse shapes and behaviors, even in a limited 2-D design space.

In the rest of this section, we investigate where the performance difference arises during the co-optimization with different controllers. In particular, we investigate whether modular control has a better inductive bias or transfers better to similar morphologies.

### 4.2 Optimization on fixed morphologies

Similar to [17, 22, 26, 34, 43], we optimize both controllers for heuristically chosen fixed morphologies to see whether modular control has a better inductive bias. Figure 5 shows the experimented fixed morphologies. Biped (Fig. 5a) and worm (Fig. 5b) are commonly experimented morphologies in the literature [17, 34, 37, 43]. We also experiment with less commonly used morphologies such as triped (Fig. 5c) and block (Fig. 5d) as we intuitively think that they require different locomotion strategies. For these experiments, we only considered morphologies made out of a single material for ease of design. We optimize global and modular controllers for each morphology for 1500 generations, repeat the experiments with different random seeds 10 times, and report 95% confidence intervals.

The results are illustrated in Figure 6, where we see the fitness of the best individual at each generation. Interestingly, modular controllers' advantages in the co-optimization setting are not observed when controllers are directly optimized for a fixed morphology. Both controllers achieve comparable performances for Worm, Triped, and Block ($p > 0.05$), and the global controller performs better on Biped ($p < 0.01$). Moreover, we don't observe faster convergence of modular controllers. While both types of controllers are evolved to locomote the robots successfully, the performance on certain morphologies, such as Biped, is better than other experimented morphologies. Since both controllers have similar performances when optimized for a single morphology, we hypothesize that the performance difference between the two control paradigms during the co-optimization arises from their effects on the search over the morphology space. This agrees with earlier findings showing that the key to successful brain-body co-optimization is preventing premature convergence of body plan via fragile co-adaptation between a robot's controller and morphology [6, 7].

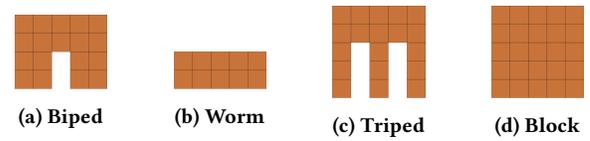

(a) Biped　(b) Worm　(c) Triped　(d) Block

Figure 5: Experimented fixed morphologies.

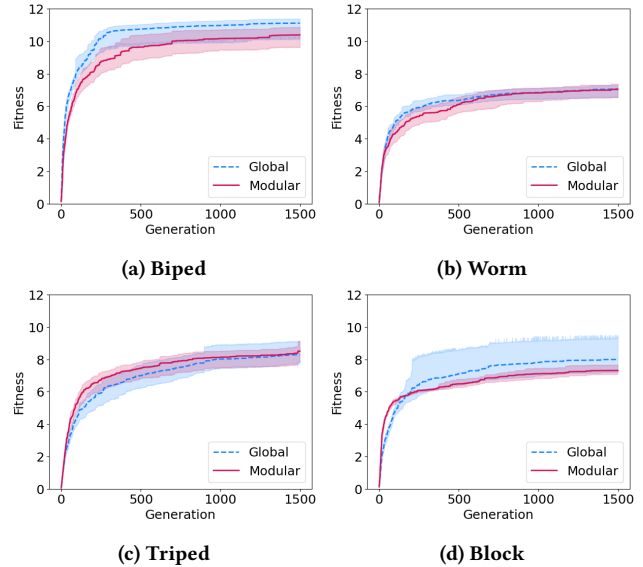

(a) Biped　(b) Worm

(c) Triped　(d) Block

Figure 6: Fitness of the best individual over evolutionary time for fixed morphology experiments. Modular controllers do not perform better than global controllers when trained in isolation on any of these single morphologies.

So the question is this: how can modular controllers prevent prematurely eliminating underperforming yet promising body plans from the population? Ideally, we would like to assess the fitness potential of a new body plan correctly, which can be approximately achieved by reducing the detrimental effects of morphological changes on immediate fitness. To test whether modular controllers have an advantage in this sense, we investigate their ability to control multiple morphologies and transferability to other morphologies.

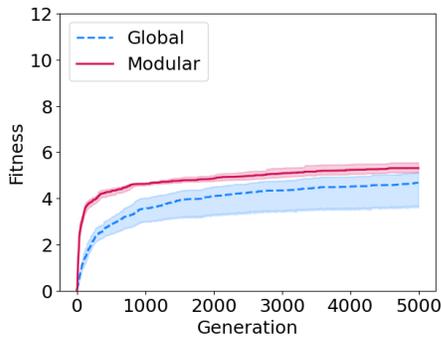

**Figure 7: Fitness over evolutionary time plot for joint training on all four of the fixed morphologies. Even though end performances are comparable, modular control converges quicker than global control, suggesting ease of optimization for different morphologies simultaneously. We omit the performance trajectories on individual morphologies as they follow the joint fitness very closely.**

## 4.3 Optimization on multiple fixed morphologies

If a controller can successfully control multiple morphologies, it can be transferred from one morphology to another without drastic performance drops. Even though the controllers are not explicitly optimized for multiple morphologies during the co-optimization of the brain and body, we experiment with joint training on multiple fixed morphologies to determine their potential for this task.

Similar to [22], we optimize both controllers for all experimented fixed morphologies (see Figure 5) jointly. Inspired by [45], we use the minimum performance among all morphologies as the fitness values for the controllers. 10 trials for each type of controller, initialized with different random seeds, are evolved for 5000 generations.

As shown in Figure 7, both controllers achieve comparable performances ($p > 0.05$), yet the populations with modular controllers converge more quickly. This demonstrates the advantage of modular control over global control; modular controllers can be optimized for multiple morphologies significantly faster. While the co-optimization setting doesn't explicitly optimize controllers for multiple morphologies, adapting quickly to multiple morphologies is important for quickly recovering from the detrimental effects of morphological changes. Combined with our evolutionary algorithm that optimizes for age and fitness where new individuals face less selection pressure, individuals with modular controllers may have a better chance of demonstrating their potential.

Additionally, we note that the performances on multiple morphologies are lower than on single morphologies, suggesting that performance and generalization are competing objectives. Nonetheless, successful locomotion behaviors are evolved. This is in contrast to [22], where joint training on multiple rigid fixed robots oftentimes performs very poorly. We conjecture that this may be due to soft robots' compliance. While rigid robots require distinct strategies to locomote, soft robots' compliance allows for less effective yet functional strategies that work on multiple different soft robots.

## 4.4 Transferability to other morphologies

To further support our claims, we check controllers' transfer performances to similar morphologies. To create similar morphologies, we mutate the original morphologies and create *neighboring morphologies in the mutation space.* To see how the controller's transfer performance is affected by the (dis)similarity of the new body plan to the original morphology, we also sample neighbors with different distances, where distance is the number of mutations applied.

Given the way of creating similar morphologies, we start with controllers optimized for single fixed morphologies. The champion of each run from Section 4.2 is transferred to morphologies some number of mutations away from the body plan that they are optimized for. 20 distinct neighboring morphologies are sampled per distance, and controllers' zero-shot (no additional training) and one-shot transfer (one generation of controller evolution ) performances, as the relative change in fitness, are measured.

Figure 8 illustrates the results of transferring controllers optimized for a single morphology to the neighbors of that morphology. Robots with both controllers' performance drop as they are transferred to more dissimilar morphologies. We also see that both controllers regain performance when as little as one generation of controller finetuning is performed on the new morphology (one-shot transfer), suggesting the ability to rapidly re-adapt controllers to near morphologies. The finding that modular controllers' transfer performance is never worse and oftentimes better than the global controllers (for both zero and one-shot transfer), even though both perform similarly when optimized for a single morphology, suggests that modular controllers are not inherently better at finding effective locomotion strategies for soft robots, but that they are significantly more robust and adaptable to morphological changes.

The above findings apply to transferred controllers that are optimized for a particular fixed morphology for 1500 generations. This doesn't necessarily represent the issues of fragile co-adaptation that may occur during a robot's brain-body co-optimization process. To convincingly demonstrate the transferability of modular controllers during co-optimization, we take the best individuals from different generations during the co-optimization process and measure their zero-shot and one-shot transfer performances.

There is one difference with the previous transferability experiment. Each individual sampled from co-optimizations runs potentially has a different morphology. Therefore in this experiment, each sampled controller is transferred to a potentially different set of neighbors obtained by mutating their original body.

Figure 9 demonstrates the transferability of controllers sampled from co-optimization runs. The transfer performance of modular control is never worse and oftentimes significantly better compared to global control. The performances drop as the distance to the original morphology increases. Both trends are consistent in zero and one-shot settings and all sampled generations.

Modular controllers' ability to transfer better helps during co-optimization by increasing the chances of survival for individuals with newly modified morphologies. If a controller transfers better to a similar morphology, the time required to adapt the controller fully will be shorter, and the probability of eliminating a promising morphology from the population will be lower. This results in a better search over the morphology space, even without methods aimed at protecting new morphologies during search [7]. Figure 10 illustrates the advantage that modular controllers confer to morphological search – enabling changes to the morphology to make up a significantly larger percentage of the successful mutations that enable brain-body co-optimization ($p < 0.05$). This is true both for the successful mutations that eventually lead to the champions of each run (10a) and for all successful mutations to any individual throughout the search process (10b).

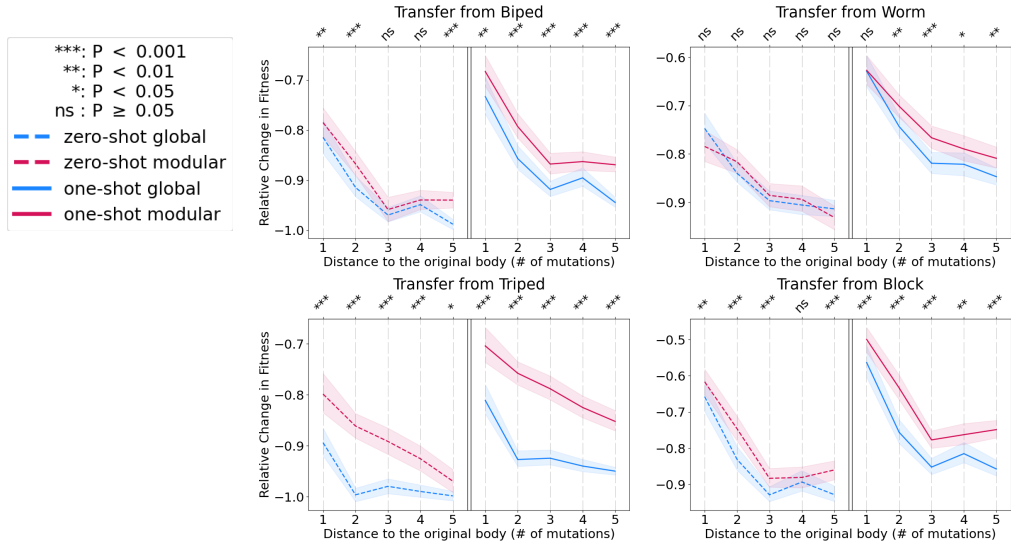

Figure 8: Zero-shot transfer (performance immediately after a morphological mutation; dotted lines on the left side) and one-shot transfer (after one epoch of controller evolution; solid lines on the right side) performance of controllers optimized for fixed morphologies. The champion of each run is transferred to neighboring morphologies with increasing dissimilarity (number of mutations away on the x-axis). All settings show an average decrease in performance upon morphological transfer (negative relative change on the y-axis), but the modular control is never worse and often significantly outperforms (top axis) the global controller at transfer to morphological neighbors.

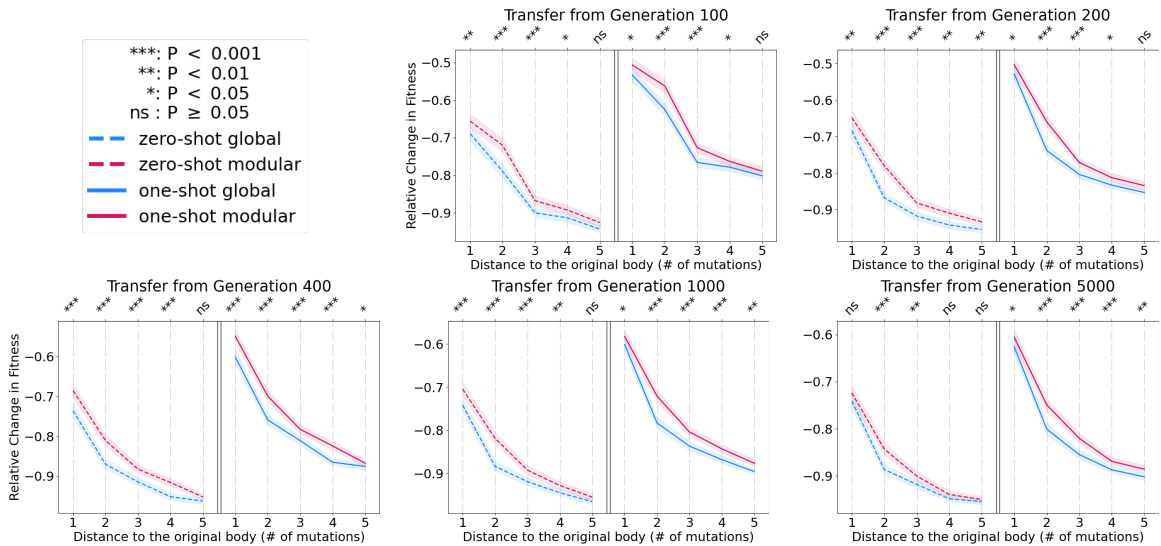

Figure 9: Performance of controllers sampled from various points in evolutionary time for brain-body co-optimization when transferred to their morphological neighbors of various distances (See Fig. 8 for walk-through of figure components). Despite consistently negative morphological mutations, modular control is never worse and often more robust/adaptable, with significantly smaller drops in fitness compared to global control for both zero and one-shot morphological transfer.

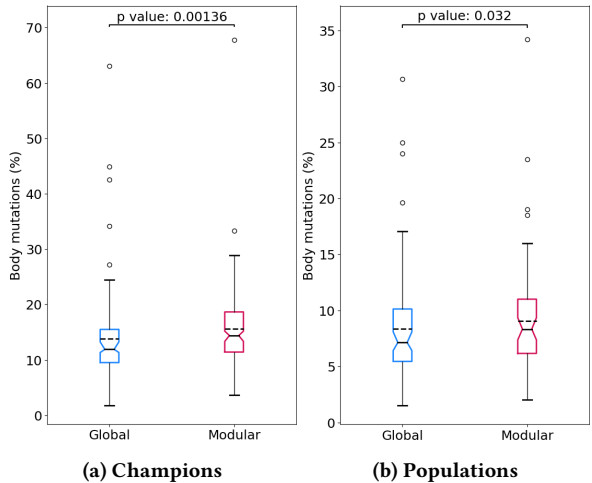

(a) Champions  (b) Populations

Figure 10: Boxplots showing the proportion of successful mutations during co-optimization that modifies the body plan of the robot (as opposed to its controller). Robots evolved with modular controllers attribute a larger proportion of their successful offspring to modifications to their body plans than robots with global controllers do. This is true for lineages leading up to the run champions (a) and for all positive mutations to the whole population throughout evolution (b).

## 5 DISCUSSION

The findings above, that robots undergoing brain-body co-optimization with modular controllers lead to a greater number of successful morphological mutations and higher overall fitness values than those employing global controllers, suggest that enabling successful morphological mutations is a key to enabling successful brain-body co-optimization in evolving soft robots. Our demonstration that modular controllers more effectively transfer to children produced by morphological mutations helps to provide evidence for a mechanistic understanding of how these controllers convey that advantage within the evolutionary co-optimization process. The adaptability and robustness of modular controllers, relative to their similarly sized but globally oriented counterparts, may not be entirely surprising and is supported by general intuition and prior findings [30, 63]. But we are not aware of previous work that has demonstrated the connection between these more adaptable controllers and the importance of being adaptable/robust to morphological mutations during the evolutionary co-optimization process.

This work investigates the effect of controller robustness to morphological perturbations. Inspired by the positive findings here, future work will similarly investigate the effect of morphologies which are more robust to controller perturbations. Their robustness may be due to the implementation choices of various types or encodings of morphologies (as done here for controllers), but the optimization of bodies and brains which are evolved or trained specifically to be robust, rapidly adaptable [18], or to recover in few shots [62] will be of great interest and value. Thus, this work ties into the broader study of the evolution of robustness [1, 4, 16, 23] and how robustness affects the evolution of evolvability [2, 15, 59].

The potential for rapid re-adaptation of controllers to similar morphologies demonstrated in Figure 8 also calls into question the overarching perspective and impetus of this work given in the introduction (that we are focusing here on exploring methods for avoiding fragile co-adaptation), as controllers can gain back a significant portion of the performance lost during the morphological mutation with just one generation of retraining (one-shot performance on both controllers is significantly higher than zero-shot performance with no re-adaptation). However, the finding that even with a one-shot update for re-adaption, the robot's performance is still less than half of what it was prior to the morphological mutation supports the idea that fragile co-adaptation is a serious issue and investigating methods to avoid will likely be of value is in the presence of more advances rapid re-adaption strategies.

This presence of negative morphological mutations was touted as a major hurdle for brain-body co-optimization in [6] and methods to sidestep were presented explicitly in [7] and implicitly in [31]. While we do not use such methods that specifically look for diversity of new morphologies here, our use of an evolutionary algorithm highly incentivizing age-based diversity (Sec. 3.4) may be an important part of maintaining search despite these negative mutations. Additional future work is ongoing to investigate this.

In this work, we explored an elementary version of modular control in a relatively small 2D design space and utilized a simple locomotion task for performance evaluation. In future works, complex control strategies in more complicated 3D design spaces and harder tasks involving more environmental observations and closed-loop information processing should be investigated, as we believe these settings present greater challenges for co-optimization and will inform the general applicability of this approach. Moreover, the advantages presented in the paper are significant but small. As the implementation decision to employ modular control is agnostic to the evolutionary algorithm it is paired with, combining this approach with algorithms specifically designed to aid brain-body co-optimization, such as [7], can be easily explored and may synergistically enhance co-optimization further.

## 6 CONCLUSION

In this paper, we investigate the potential of modular control of soft robots in the challenging co-optimization setting where both the morphology and the control are optimized together. We show that modular control enjoys better co-optimization performance in this setting. It converges faster and finds better solutions. Moreover, our work suggests that the performance gain arises from the better transferability of modular controllers to similar morphologies, enabling efficient search over morphology space. This is in line with the previous findings [7] and the theory of embodied cognition.

## ACKNOWLEDGMENTS

This material is based upon work supported by the National Science Foundation under Grant No. 2008413. Computations were performed on the Vermont Advanced Computing Core supported in part by NSF Award No. OAC-1827314.